\documentclass{article}

\usepackage{microtype}
\usepackage{graphicx}
\usepackage{subfigure}
\usepackage{booktabs} 

\usepackage{hyperref}

\usepackage[accepted]{icml2025}

\usepackage{amsmath}
\usepackage{amssymb}
\usepackage{mathtools}
\usepackage{amsthm}

\usepackage[capitalize,noabbrev]{cleveref}

\theoremstyle{plain}

\theoremstyle{definition}

\theoremstyle{remark}

\usepackage[textsize=tiny]{todonotes}

\icmltitlerunning{Scalpel vs. Hammer: GRPO Amplifies Existing Capabilities, SFT Replaces Them}

\begin{document}

\twocolumn[
\icmltitle{Scalpel vs. Hammer: GRPO Amplifies Existing Capabilities, SFT Replaces Them}



\icmlsetsymbol{equal}{*}

\begin{icmlauthorlist}
\icmlauthor{Neel Rajani}{edi}
\icmlauthor{Aryo Pradipta Gema}{edi}
\icmlauthor{Seraphina Goldfarb-Tarrant}{cohere}
\icmlauthor{Ivan Titov}{edi,ams}
\end{icmlauthorlist}

\icmlaffiliation{edi}{Institute for Language, Cognition and Computation (ILCC), University of Edinburgh, United Kingdom}
\icmlaffiliation{cohere}{Cohere, London, United Kingdom}
\icmlaffiliation{ams}{Institute for Logic, Language and Computation (ILLC), University of Amsterdam, Netherlands}
\icmlcorrespondingauthor{Neel Rajani}{Neel(dot)Rajani(at)ed.ac.uk}

\icmlkeywords{Machine Learning, ICML}

\vskip 0.3in
]



\printAffiliationsAndNotice{}  

\begin{abstract}
\looseness-1
Training large language models (LLMs) for reasoning via maths and code datasets has become a major new focus in LLM post-training. Two particularly popular approaches are reinforcement learning (RL) and supervised fine-tuning (SFT), but their training dynamics are poorly understood. We present a comparative analysis of RL and SFT on the same maths problems with the same model and similar hyperparameters. We find that RL yields minor in-domain gains on maths and slight degradation on knowledge-intensive benchmarks like MMLU, while both trends are more pronounced in SFT. We also analyse model parameters across checkpoints, observing that both algorithms modify query and key weights the most. Meanwhile, SFT exhibits greater updates and also affects mid-layer MLPs more, leading us to hypothesise that this may have caused the out-of-domain degradation. We therefore investigate whether freezing parts of the model during training can mitigate the reduced performance on knowledge-intensive benchmarks. However, our results are inconclusive, with benefits on GPQA:Diamond and degradation on other benchmarks. Taken together, our observations provide a preliminary indication for why RL amplifies existing capabilities, while SFT replaces old skills with new ones.
\end{abstract}

\section{Introduction}
\label{sec:introduction}
Large language model (LLM) research has seen the emergence of a new post-training paradigm. When OpenAI released o1 \cite{openai2024openaio1card}, it triggered a wave of interest in ``reasoning models''. These are models trained to generate chain-of-thought (CoT) traces in which they break down problems, work through them step by step and verify intermediate results. Remarkably, this causes them to be more reliable in a wide range of domains, and has advanced the frontier of performance as measured by standard benchmarks. ``Reasoning training'' has become very popular, and is performed with reinforcement learning (RL) algorithms like GRPO, or with supervised fine-tuning (SFT) on synthetic datasets collected from other reasoning models \cite{deepseek2025deepseekr1}. However, there is little understanding of the dynamics that underpin these algorithms, and how they affect models internally. Furthermore, recent frontier models trained for reasoning with RL have been shown to hallucinate more \cite{o34syscards}. It is paramount for the field to gain a better understanding of how reasoning training affects models.

As such, we present a comparative analysis of GRPO and SFT under a highly controlled setup. We design our experiments to eliminate as many confounding variables as possible. We first apply GRPO to open-source models (Section \ref{sec:reasoning_training_with_GRPO}), finding it to be expensive and unstable, requiring many tweaks for training to be successful. To enable fair comparisons, we then adapt our most performant hyperparameter configuration to SFT as faithfully as possible (Section \ref{sec:reasoning_training_with_SFT}), observing training to be much more stable and reliable. We then compare both approaches across a diverse set of benchmarks (Section \ref{sec:evaluating_models_after_reasoning_training}), and note a trade-off between the two: SFT gives rise to more pronounced gains in-distribution at the cost of capability degradation out-of-domain. Meanwhile, performance improvements are more modest for GRPO in return for out-of-domain capabilities staying more intact. 

We then ask how GRPO and SFT affect models differently across checkpoints. To answer this question, we turn towards 20 intermediate checkpoints saved during training for each algorithm. We compute the KL divergence of every checkpoint with the base model (Section \ref{sec:KL_divergence_throughout_training}), observing SFT to have a particularly pronounced impact early into training, while GRPO impacts models more gradually. We also compare models on a parameter level (Section \ref{sec:cross-checkpoint_analysis_of_reasoning_training}), observing that both approaches adjust query and key weights more than other weights. Notably, we find that SFT causes much greater updates than GRPO. This is especially pronounced in middle layers, which have been found to be involved in memorising factual associations \cite{geva2021transformerfeedforwardlayerskeyvalue, meng2023locatingeditingfactualassociations}. We hypothesise that this may be the reason for SFT's greater degradation on knowledge-intensive benchmarks.

This motivates us to further investigate whether we can apply these insights to adjust training. We ask if freezing the MLP matrices can preserve factual knowledge. While we find some evidence of benefits in-domain and out-of-domain, our results are inconclusive. Further, we investigate whether training only the query and key matrices during SFT is sufficient, but find this approach to cause unchanged or reduced benchmark performance. 

Taken together,  our results provide preliminary indications for how GRPO and SFT affect models differently. While our findings require further validation, they offer an initial glimpse into the training dynamics that underpin the effects of GRPO and SFT on reasoning.

\section{Related works}
\label{sec:related_works}
\subsection{Reasoning-focused training}
\label{sec:reasoning-focused_training}
\looseness-1
Since the release of O1 \cite{openai2024openaio1card}, there has been a flurry of new ``reasoning'' models\footnote{Whether these models are genuinely ``reasoning'' is a question we leave for other work. We use the term to refer to models which perform better on a wide range of tasks because they have been trained with RL to generate long chain-of-thought traces.
} from frontier labs \cite{Claude3.7, Gemini2.5, Grok3}. The key to the progress of ``reasoning models'' lies in applying reinforcement learning with verifiable rewards \citep[RLVR;][]{lambert2025tulu3pushingfrontiers} directly to base models. Maths and code are popular domains for this as they are comparatively easy to verify. DeepSeek's release of R1 \cite{deepseek2025deepseekr1} is particularly noteworthy as they describe the RLVR algorithm they use, Group Relative Policy Optimization \citep[GRPO;][]{shao2024deepseekmathpushinglimitsmathematical}. Furthermore, they explain their process of applying RL directly to a base model, and collecting reasoning traces from a model trained with RL to create a synthetic dataset for supervised fine-tuning (SFT). Across both methods, they observe noteworthy gains in performance.  

Their open release and prominent claims gave rise to a further wave of research into GRPO. Notably, \citet{liu2025understandingr1zeroliketrainingcritical} discover that \citet{deepseek2025deepseekr1}'s claims of 'aha moments' from GRPO may be over-exaggerated, as the base model already exhibits them. Further, they argue that longer response lengths stem from length and difficulty biases in the GRPO setup concurrently to \citet{yu2025dapoopensourcellmreinforcement}\footnote{For an approachable breakdown of various implementations and adjustments to GRPO, we refer the reader to \url{https://youtu.be/amrJDwMUFNs?si=OKNpUxesyZz9bWHQ}}.

In addition, it has been observed that capability acquisition occurs during pre-training and continual fine-tuning, while GRPO mainly amplifies skills the base model already has 
\cite{zhao2025echochamberrlposttraining, ma2025reasoningmodelseffectivethinking, ai2025rethinkingreflectionpretraining, gandhi2025cognitivebehaviorsenableselfimproving}. \citet{yue2025doesreinforcementlearningreally} show that these can be already be exposed in the base model by sampling with pass@\textit{k} for sufficiently large values of \textit{k} . Further, \citet{hochlehnert2025soberlookprogresslanguage} observe that RL may be prone to over-fitting on small benchmarks and that results are sensitive to sampling parameters and random seeds. Meanwhile, \citet{mukherjee2025reinforcementlearningfinetunessmall} find that RL makes sparse updates to subnetworks, and in concurrent work, \citet{huang2025blendingsupervisedreinforcementfinetuning} and \citet{yan2025learningreasonoffpolicyguidance} show that blending RL and SFT can mitigate limitations of either one. Taken together, these findings suggest that GRPO primarily serves capability amplification by shifting the output distribution towards correct completions which the base model was already capable of. Our work is in line with these results and comprises a first investigation into the training dynamics that underpin of this phenomenon.

\subsection{Mechanistic interpretability}
\label{sec:mechanistic_interpretabilitys}
Mechanistic interpretability seeks to reverse engineer operations performed by transformers \cite{elhage2021mathematical}, for example by identifying circuits \cite{olsson2022context, wang2022interpretabilitywildcircuitindirect, prakash2024finetuningenhancesexistingmechanisms} or features \cite{bereska2024mechanisticinterpretabilityaisafety}. It has provided promising initial insights into how LLMs work, for example by showing that mid-layer MLPs are critical for factual associations \cite{geva2021transformerfeedforwardlayerskeyvalue, meng2023locatingeditingfactualassociations}, and that induction heads drive in-context learning \cite{brown2020languagemodelsfewshotlearners, olsson2022context}. While prior work has investigated training dynamics during pre-training \cite{tigges2024llmcircuitanalysesconsistent, ai2025rethinkingreflectionpretraining, sun2025amurocharanalyzingrelationship, biderman2023pythiasuiteanalyzinglarge}, cross-checkpoint comparisons during fine-tuning remain under-researched. Our work addresses this research gap, with a specific focus on training models for reasoning.

\section{Training models to reason}
\label{sec:training_models_to_reason}
We start by exploring the differences between RL and SFT from the perspective of training effectiveness and benchmark performance. To enable principled comparisons, we line up the training setups for both algorithms as closely as possible (for more details, refer to Appendix \ref{sec:appendix:reproducibility}). We first tune GRPO and then adapt the hyperparameters to SFT. For our successful training runs, we save 20 intermediate checkpoints as this is a good trade-off between storage space and sufficient data points for subsequent analysis. We use OLMo-2-1124-7B-Instruct \cite{olmo20252olmo2furious} as it is the most capable model at the 7B scale with openly available training data, enabling subsequent investigation into when specific capabilities were acquired.

Our training data comes from the OpenR1-Math-220k dataset\footnote{\url{https://huggingface.co/datasets/open-r1/OpenR1-Math-220k}} for all training runs, because it allows for GRPO and SFT on the same questions. It consists of reasoning traces generated by DeepSeek R1 in response to questions from NuminaMath 1.5 \cite{numina_math_datasets} that were verified with Math Verify\footnote{
\url{https://github.com/huggingface/Math-Verify}
} and Llama-3.3-70B-Instruct \cite{grattafiori2024llama3herdmodels} as an LLM judge. The questions come from a variety of sources such as maths contests, K-12 math textbooks, olympiads, or forums. 

For clarity, we briefly emphasize the difference between the two approaches. During our GRPO training, the questions come from NuminaMath 1.5, and the completions are sampled from the model itself. The parameter updates depend on how the reward functions assess the completions. In contrast, during SFT the questions also come from NuminaMath 1.5, but the completions were sampled from DeepSeek R1. Parameter updates do not depend on anything beyond a model's ability to predict R1's completions given the prompt. In both setups, we set all parameters to be trainable, and use the model's native precision (\texttt{bfloat16}). Both setups also use the same prompt template, which is similar to the one used by DeepSeek \cite{deepseek2025deepseekr1}. The system prompt instructs the model to use the \texttt{\textless think\textgreater ...\textless /think\textgreater} and \texttt{\textless answer\textgreater...\textless /answer\textgreater} format. A full prompt-completion pair can be found in Figure \ref{fig:xml_tags}.

\subsection{Reasoning training with GRPO}
\label{sec:reasoning_training_with_GRPO}

\begin{figure}[h]
\begin{center}
\centerline{\includegraphics[width=\columnwidth]{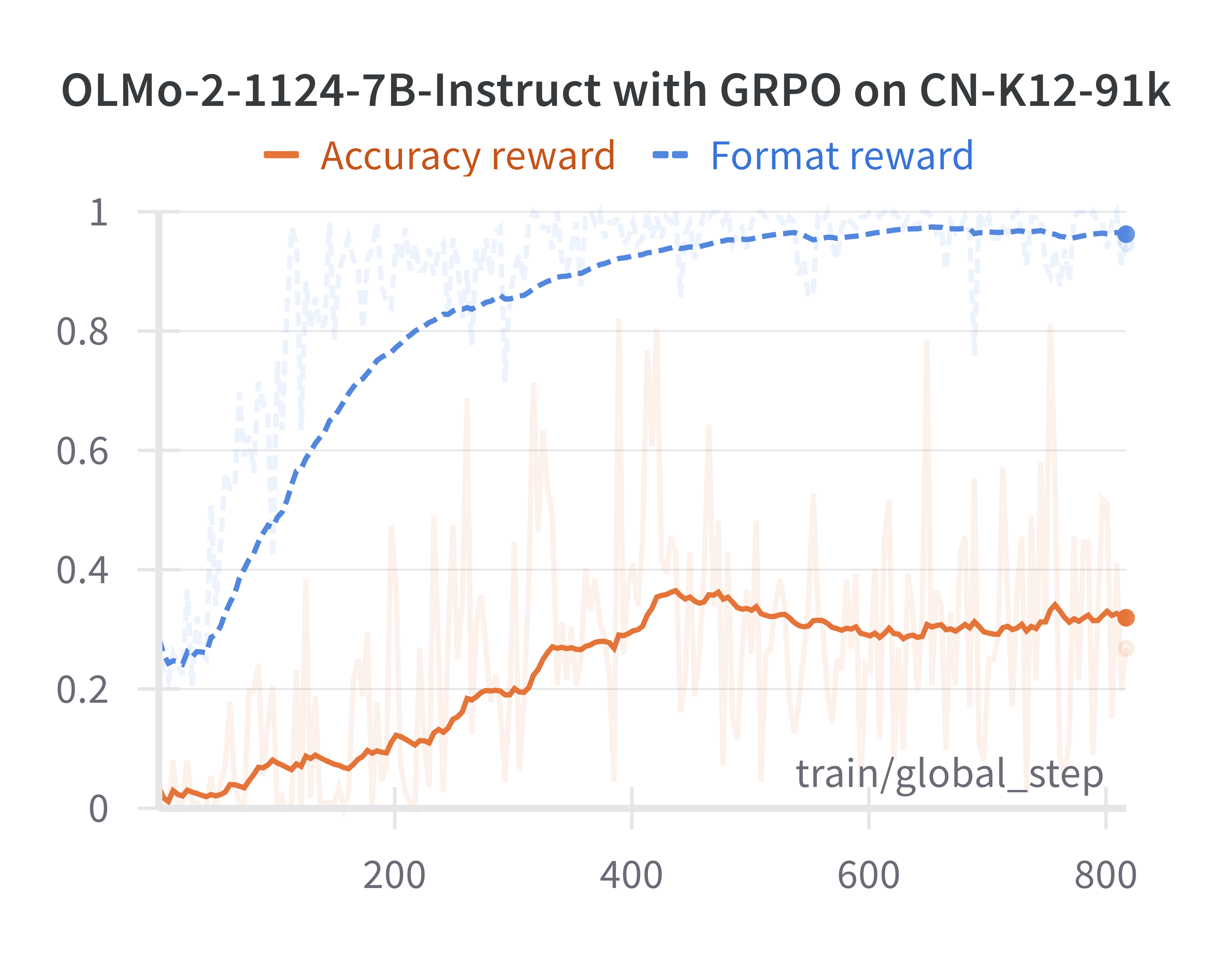}}
\caption{Accuracy and format rewards of OLMo-2-1124-Instruct during GRPO on questions sampled from CN-K12-91k subset. Format rewards increase quickly as the model follows instructions correctly, while accuracy rewards increase more slowly and plateau.}
\label{fig:format_accuracy}
\end{center}
\vskip -0.1in
\end{figure}

We successfully train open-source models on maths tasks with GRPO, though we observe it to be computationally expensive and unstable. In our efforts to find effective training configurations, we develop a number of critical insights. Firstly, we find that many questions in OpenR1-Math-220k are at the olympiad level and too challenging for OLMo-2-1124-7B-Instruct to answer correctly. This results in many parameter updates with respect to incorrect reasoning traces. Moving instead to an easier subset of 91k K12-level questions yielded much more reliable results, with accuracy and format rewards shown in Figure \ref{fig:format_accuracy} (refer to Appendix \ref{sec:appendix:question_difficulty} for further details). Secondly, we note that careful reward function design is critical for rewarding responses that are both accurate and correctly formatted. We find reliable results when scaling the accuracy rewards and the format rewards by 1.0 and 0.2, respectively (refer to Appendix \ref{sec:appendix:reward_function_design} for further details). Finally, we make many additional adjustments to training hyperparameters to improve training stability (refer to Appendix \ref{sec:appendix:training_stability} for further details). We list the hyperparameters of our final approach and all details for reproducibility in Appendix \ref{sec:appendix:reproducibility}, and insights from attempts and recipes that worked poorly are provided in Appendix \ref{sec:appendix:GRPO_failures}.

\subsection{Reasoning training with SFT}
\label{sec:reasoning_training_with_SFT}
\begin{figure}[h]

\begin{center}
\centerline{\includegraphics[width=\columnwidth]{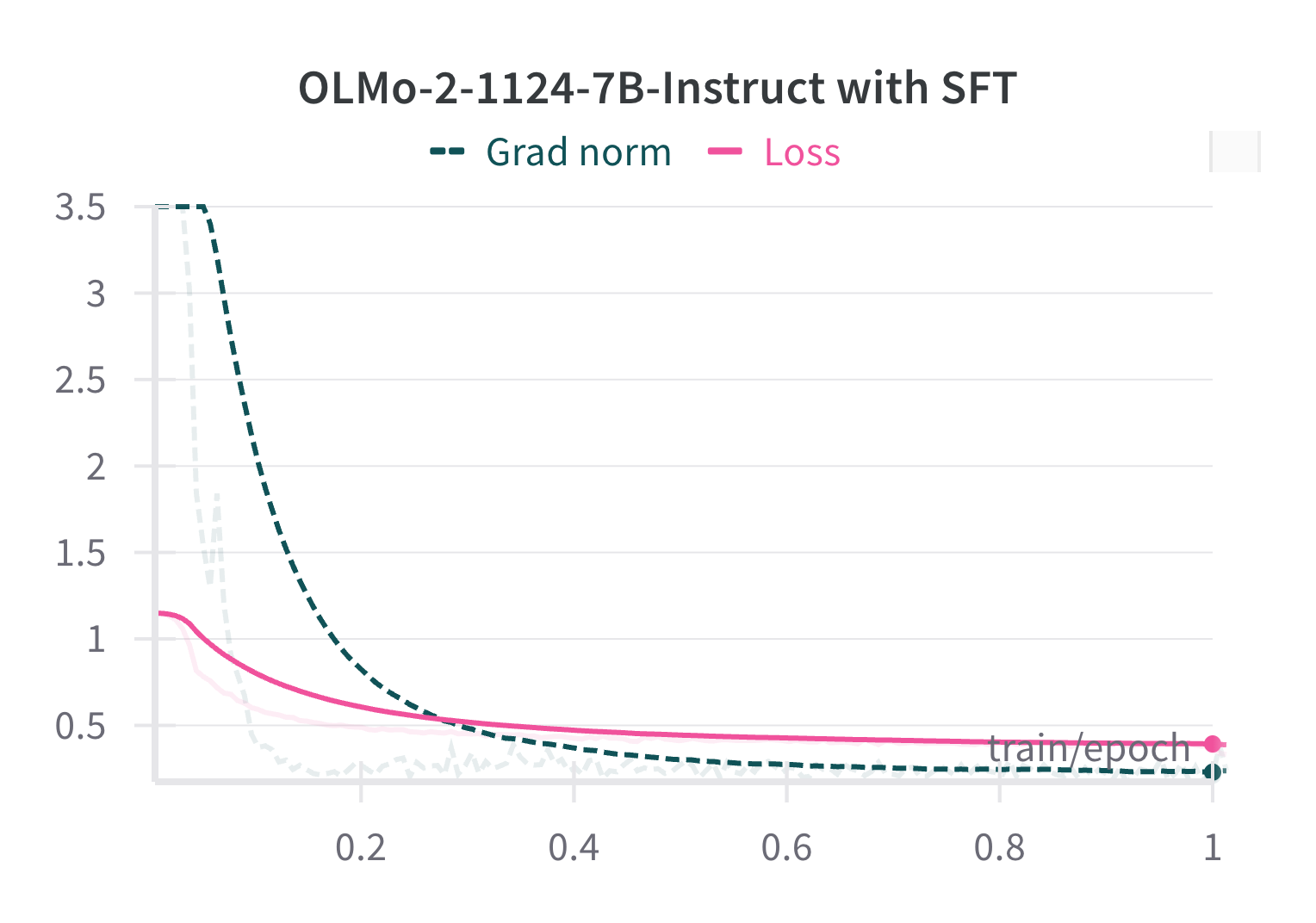}}
\caption{The gradient norm and loss curves of OLMo-2-1124-Instruct during SFT on questions and completions from the CN-K12-91k subset.}
\label{fig:grad_loss}
\end{center}
\vskip -0.1in
\end{figure}

\begin{figure*}[h]
\begin{center}
\centerline{\includegraphics[width=\textwidth]{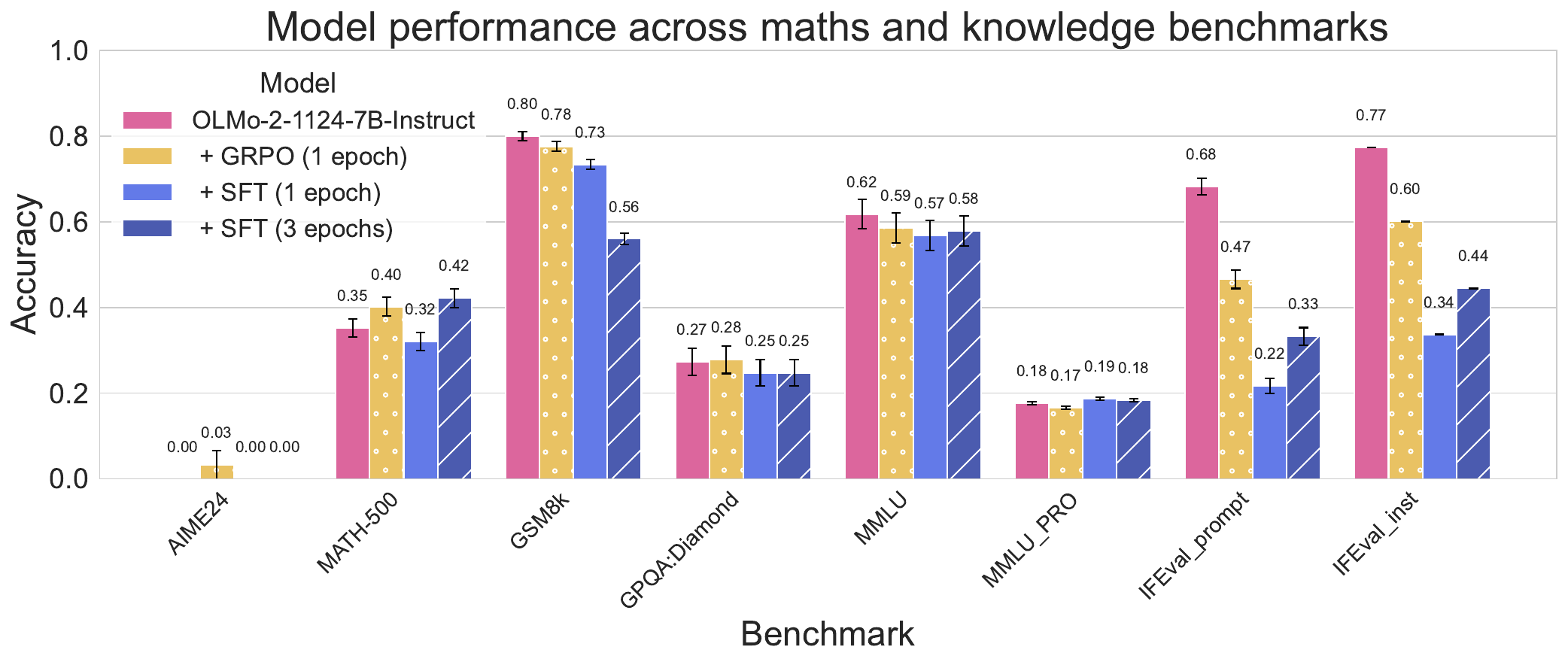}}
\caption{Evaluation results on a wide range of benchmarks. We produce all results with extractive match, and reuse sampling parameters from training (same system prompt, temperature=0.7 and top\_p=0.95). Unless otherwise indicated, all benchmarks are zero-shot CoT.}
\label{fig:benchmarks}
\end{center}
\vskip -0.1in
\end{figure*}

Following our experiments with GRPO, we adopt the working configuration from GRPO to SFT. We make a best effort attempt at keeping as many hyperparameters constant as possible, including the learning rate scheduling, the total number of questions seen by the model, the batch sizes, and the random seed. We expectedly find SFT to be much more stable. It requires much less tuning, featuring smooth loss curves, rare or no gradient norm spikes and predictable results, as shown in our final training run in Figure \ref{fig:grad_loss}.

\looseness-1
There were some settings we were unable to align exactly. Some hyperparameters are specific to GRPO and cannot be adopted, such as the number of generations per prompt, the reward weightings or the constants $\epsilon$ and $\beta$. A notable divergence in our setups was the learning rate itself. On-policy algorithms such as GRPO have often been found to require low learning rates to prevent instability. In contrast, SFT exhibits reduced benefits from training unless relatively higher learning rates are used. We find GRPO to work the most reliably with learning rates around 1e-6, while SFT required learning rates around 5e-5 to be successful. This is clearly an important confound, as different learning rates across many hundred parameter updates can cause vastly different learning trajectories. In an attempt to mitigate this, we experiment with raising the learning rate in GRPO while increasing regularization constraints. However, we find that training remains unstable and downstream performance is lower than the base model, as delineated in Appendix \ref{sec:appendix:GRPO_failures}. Therefore, we choose to use learning rates that elicit the highest performance in each training algorithm for a realistic comparison, emphasizing that this is a limitation of our analysis.

\subsection{Evaluating models after reasoning training}
\label{sec:evaluating_models_after_reasoning_training}

Following our training runs, we ask the question whether GRPO and SFT have different impacts on model capabilities, and whether they vary across domains that are in- or out-of-distribution. In order to answer it, we first evaluate our trained models on a row of maths benchmarks to verify whether reasoning training actually improved performance. We follow the open-r1 repository in evaluating models on AIME24 \cite{aime24} and MATH-500 \cite{lightman2023letsverifystepstep}. Because of its wide-spread use, we choose to additionally evaluate on GSM8k \cite{cobbe2021gsm8k}. Contrastingly, we are also interested in whether reasoning training impacts performance on benchmarks which require wide ranges of factual knowledge and recall instead of logical reasoning or inference. If it does, this raises questions about how to balance trade-offs between the two. As such, we evaluate on GPQA-Diamond \cite{rein2023gpqagraduatelevelgoogleproofqa}, MMLU \cite{hendrycks2021measuringmassivemultitasklanguage}, MMLU-Pro \cite{wang2024mmluprorobustchallengingmultitask} and IFEval \cite{zhou2023instructionfollowingevaluationlargelanguage}. We use LightEval \cite{lighteval} for all of the benchmarks mentioned above, except for MMLU-Pro which is only supported on EleutherAI's Language Model Evaluation Harness \cite{wang2024mmluprorobustchallengingmultitask}. Note that our results for the base model differ to those by \citet{olmo20252olmo2furious}, which likely stems from the difference in evaluation pipeline. Our results are shown in Figure \ref{fig:benchmarks}.

Our benchmarks reveal a varied pattern of results. For the maths benchmarks, we observe that training the model with GRPO leads to modest performance improvements on MATH-500 and a slight degradation on GSM8k. Meanwhile, both of these developments are more pronounced after 3 epochs SFT, with slightly larger gains on MATH-500 and larger drops in performance on GSM8k. On AIME24, we observe that both the base model and all of our fine-tuned models struggle to get any questions right. Due to its relative cheapness, we also investigate training with SFT for 3 epochs. We find that this exacerbates the trend of 1 epoch training further. This is in line with findings by the open-r1 community effort, who release models trained for 3 epochs with SFT \cite{openr1update2}. 

Our results are somewhat surprising, especially with respect to GSM8k. In preliminary testing, we found 0-shot results to be very poor, with a 5-shot setting critical to elicit measurable performance. \citet{olmo20252olmo2furious} mention that they make efforts not to train on the GSM8k test set, but they do train on the GSM8k training set. In contrast, our SFT dataset consists entirely of CoT \cite{wei2023chainofthoughtpromptingelicitsreasoning} traces instead of few-shot examples. It is therefore possible that SFT corrupted the already well-optimised few-shot learning abilities of the model on this benchmark. Meanwhile, on the knowledge-intensive benchmarks that we tested with, we find that SFT generally degrades performance quite noticeably while GRPO has less of an impact. This trend is reversed in GPQA-Diamond, but the benchmark also includes questions where reasoning training might help, in domains such as physics. 

Taken together, our results suggest that reasoning training via GRPO may offer modest gains in targeted domains, while causing less degradation in others.  Meanwhile, it appears that SFT can have much greater in-domain benefit at much higher out-of-domain costs. However, this interpretation may be affected by limitations of our setup.  For example, it is possible that our parameters were not optimally tuned, so we refrain from making overly strong claims based on our results. We also note that models trained by \citet{deepseek2025deepseekr1} with SFT exhibit less capability degradation, and hypothesise this may be due to their use of 800k heavily curated samples, of which 600k were reasoning traces and 200k related to other domains. We call for more research into this trade-off.

\section{Cross-checkpoint analysis of reasoning training}
\label{sec:cross-checkpoint_analysis_of_reasoning_training}
In this section, we examine how GRPO and SFT differ in their impact on the model throughout training. Specifically, we turn towards the 20 intermediate checkpoints we saved for both algorithms in Section \ref{sec:training_models_to_reason}. We investigate training dynamics by measuring the KL divergence (Section \ref{sec:KL_divergence_throughout_training}), and by computing differences at the parameter level (Section \ref{sec:norm_of_differences_throughout_training}). Together, our results provide an early indication that GRPO makes fewer integral changes to the model and reinforces existing capabilities, while SFT leads to a more substantial departure from the base model and larger changes to the parameters.

\subsection{KL Divergence throughout training}
\label{sec:KL_divergence_throughout_training}
\begin{figure}[h]
\begin{center}
\centerline{\includegraphics[width=\columnwidth]{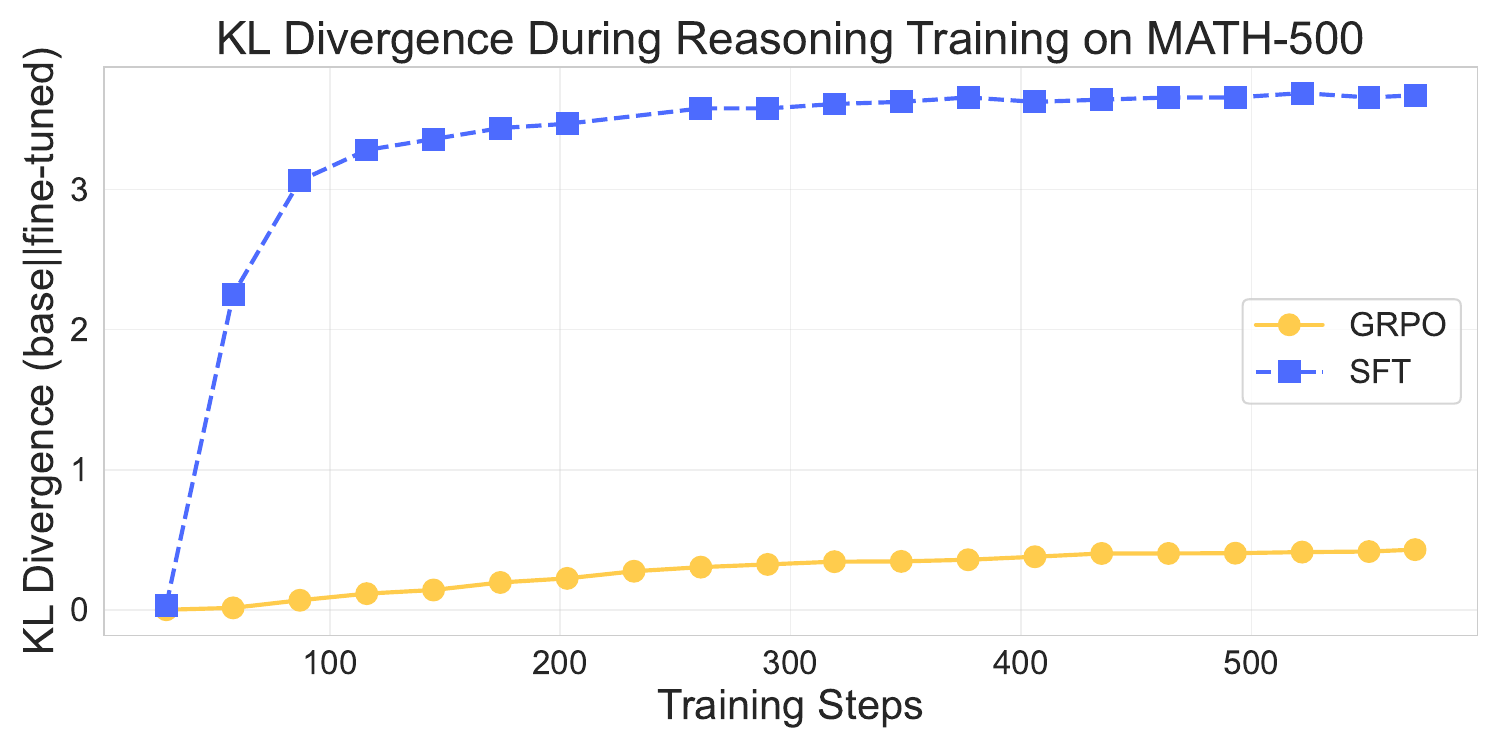}}
\caption{The KL divergence between OLMo-2-1124-7B-Instruct before training and during training on the withheld MATH-500 dataset. We find SFT to cause divergence from the base model much more quickly and to a more pronounced degree than GRPO.}
\label{fig:kl_divergence_grpo_vs_sft}
\end{center}
\vskip -0.1in
\end{figure}

Informed by the differences in benchmark scores, we further seek to complement our insights into how GRPO and SFT affect models. A summary statistic from a benchmark can conceal nuance: two models that achieve the same performance on a benchmark may in reality be generating very different answers. Therefore, we choose to augment our findings by looking at the differences between probability distributions during training as measured by the Kullback-Leibler (KL) divergence. Specifically, we compute the average per-token KL divergence between OLMo-2-1124-7B-Instruct and checkpoints throughout training for both algorithms. We collect logits on the withheld MATH-500 dataset, with questions and solutions formatted via the chat template. Our results are shown in Figure \ref{fig:kl_divergence_grpo_vs_sft}.

In line with our other findings, we observe the KL divergence between OLMo-2-1124-7B-Instruct and checkpoints during SFT to increase considerably and early into training before plateauing. Conversely, GRPO exhibits much more gradual growth in the KL divergence with the base model, with a considerably lower plateau. This indicates that after GRPO, there is not a large difference in the tokens that the two models assign high probability to. Contrastingly, the stark rise in KL divergence in SFT suggests that the model starts producing very different output distributions early on. 

\begin{figure}[t]
\begin{center}
\centerline{\includegraphics[width=\columnwidth]{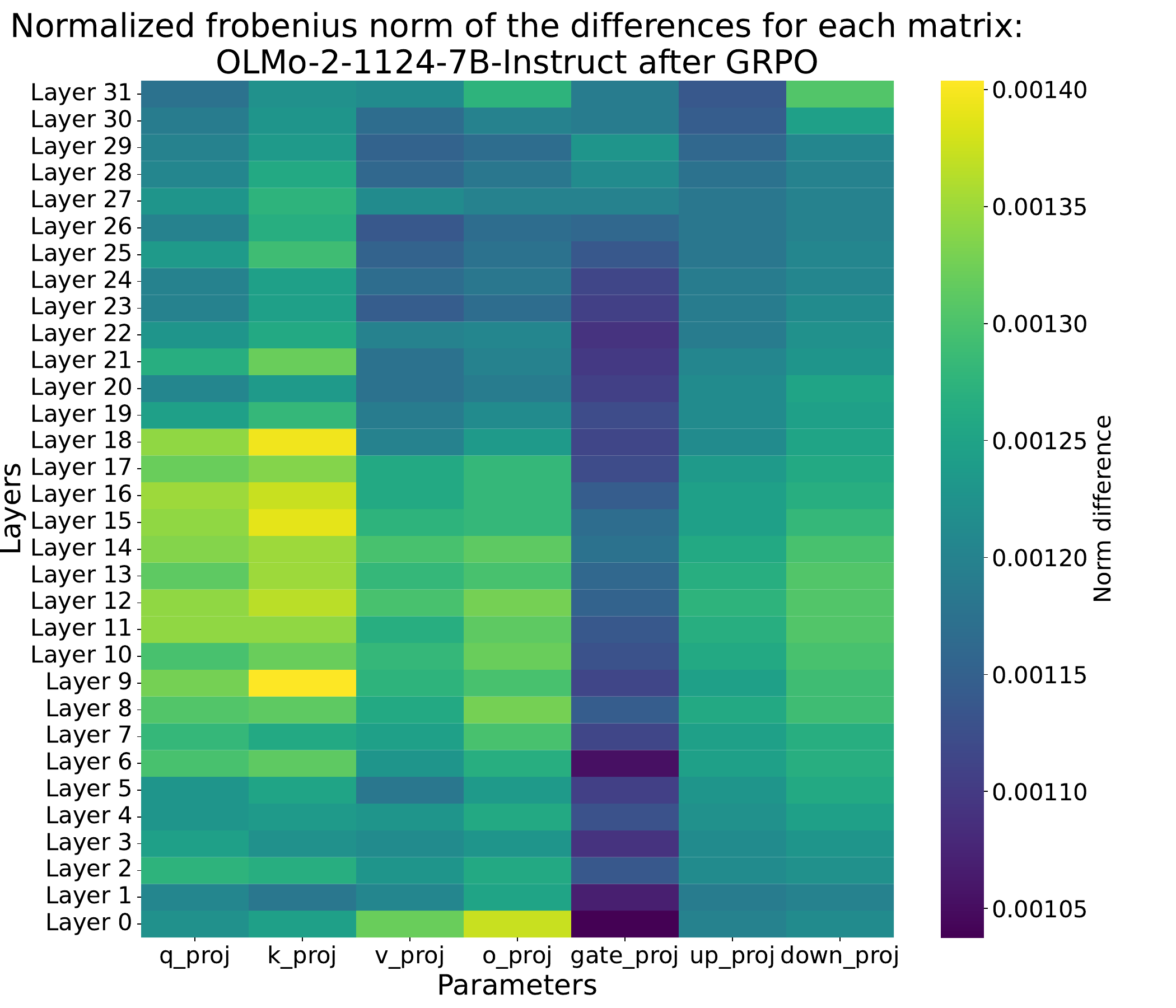}}
\caption{The parameter-level impact of GRPO, as measured by the normalized Frobenius norm of the difference between every matrix of OLMo-2-1124-7B-Instruct before and after GRPO. An animated GIF across training is available at \url{https://gifyu.com/image/bs3yP}.}
\label{fig:cckpta_grpo}
\end{center}
\vskip -0.1in
\end{figure}

These results paint a nuanced picture. On the one hand, a stronger KL divergence after SFT is expected as we use a higher learning rate. As such, updates to the model parameters are scaled more, potentially leading to more pronounced differences in output distributions. However, the early growth in KL divergence is still notable, as both models use ``constant with warmup'' for learning rate scheduling, with a warmup ratio of 0.1. This is in line with the expectedly high loss and gradient norm in this phase of training in Section \ref{sec:reasoning_training_with_SFT}, as the model still learns to fit to the tokens it is being trained on. Meanwhile, GRPO features many options for careful updates and stable training, such as clipping, which helps explain the very slow growth in KL.

\looseness-1
Fundamentally, this points back to the central difference between the two setups. In SFT, the model is taught to fit R1's CoTs, which is a very different output distribution. Intuitively, our results are therefore consistent with larger changes inside of the model and its predictions being necessary. In contrast, during GRPO, parameter updates are performed with respect to tokens that were sampled from the model itself. Again, the lower KL divergence is in accordance with parameter updates that reinforce existing capabilities instead of considerably re-organising model internals. 

\subsection{Norm of differences throughout training}
\label{sec:norm_of_differences_throughout_training}

\begin{figure}[t]
\begin{center}
\centerline{\includegraphics[width=\columnwidth]{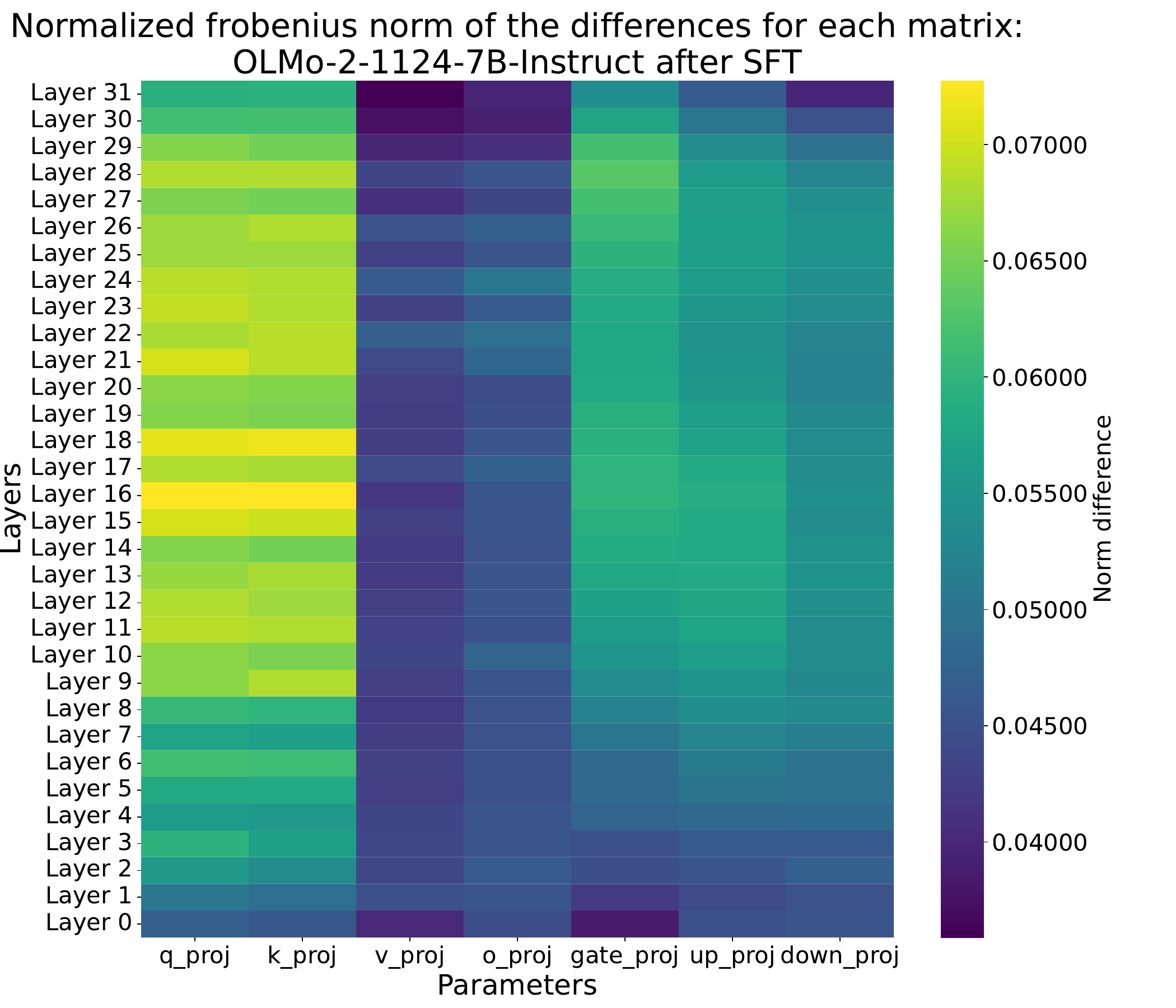}}
\caption{The parameter-level impact of SFT, as measured by the normalized Frobenius norm of the difference between every matrix of OLMo-2-1124-7B-Instruct before and after SFT. An animated GIF is available at \url{https://gifyu.com/image/bs9cB}.}
\label{fig:cckpta_sft}
\end{center}
\end{figure}

Finally, we investigate the parameter-level impact of GRPO and SFT throughout training. To this end, we take a closer look at the attention head and MLP matrices across all layers. For every checkpoint, we take the difference between the matrix in the base model and at that point in training. To be specific, this includes the four attention parameter matrices (\textbf{q}ueries, \textbf{k}eys, \textbf{v}alues and \textbf{o}utputs), and the three MLP parameter matrices (gate, up and down projections), visible on the x-axes of Figures \ref{fig:cckpta_grpo} and \ref{fig:cckpta_sft}. We then compute the normalized Frobenius norm of the difference. We exclude the query and key norm matrices (which are special to OLMo-2 \cite{olmo20252olmo2furious}), the embedding matrix, the language modelling head matrix, and the layer norm matrices in our analysis. We also emphasize a key limitation of this approach: as RL and SFT required different learning rates to achieve in-domain improvements upon the baseline, they necessarily caused different impacts internally. Therefore, we caution that the norm of differences are only a rough and approximate initial indication of which components of the model change.

\subsubsection{The parameter-level impact of GRPO}
\label{sec:param_impact_grpo}
The results of our cross-checkpoint analysis of GRPO are shown in Figure \ref{fig:cckpta_grpo}. We find that GRPO has different impacts across the layers and matrix types in OLMo-2-1124-7B-Instruct. Most notably, the largest updates seem to be concentrated in the middle-layer query and key matrices. This is interesting, as the value and output matrices of the attention heads seem to change less in comparison. In the attention mechanism, queries and keys are matrix multiplied to create the attention map. This attention map is then matrix multiplied by the values and then outputs, enabling a form of token-to-token communication. Therefore, a possible hypothesis is that GRPO causes the attention heads to attend to \textit{different} tokens, but that these still communicate the same thing. Considering the broader context of training on rollouts which included the correct solution, a potential explanation is that the backward pass causes the attention heads to focus on certain tokens differently. In maths questions, these might be central assumptions spelled out at the start of a response, or trigonometric identities while solving a sinusoidal integral. However, we emphasize that these results should be corroborated with different models and datasets before coming to stronger conclusions.

\begin{figure*}[h]
\vskip 0.1in
\begin{center}
\centerline{\includegraphics[width=\textwidth]{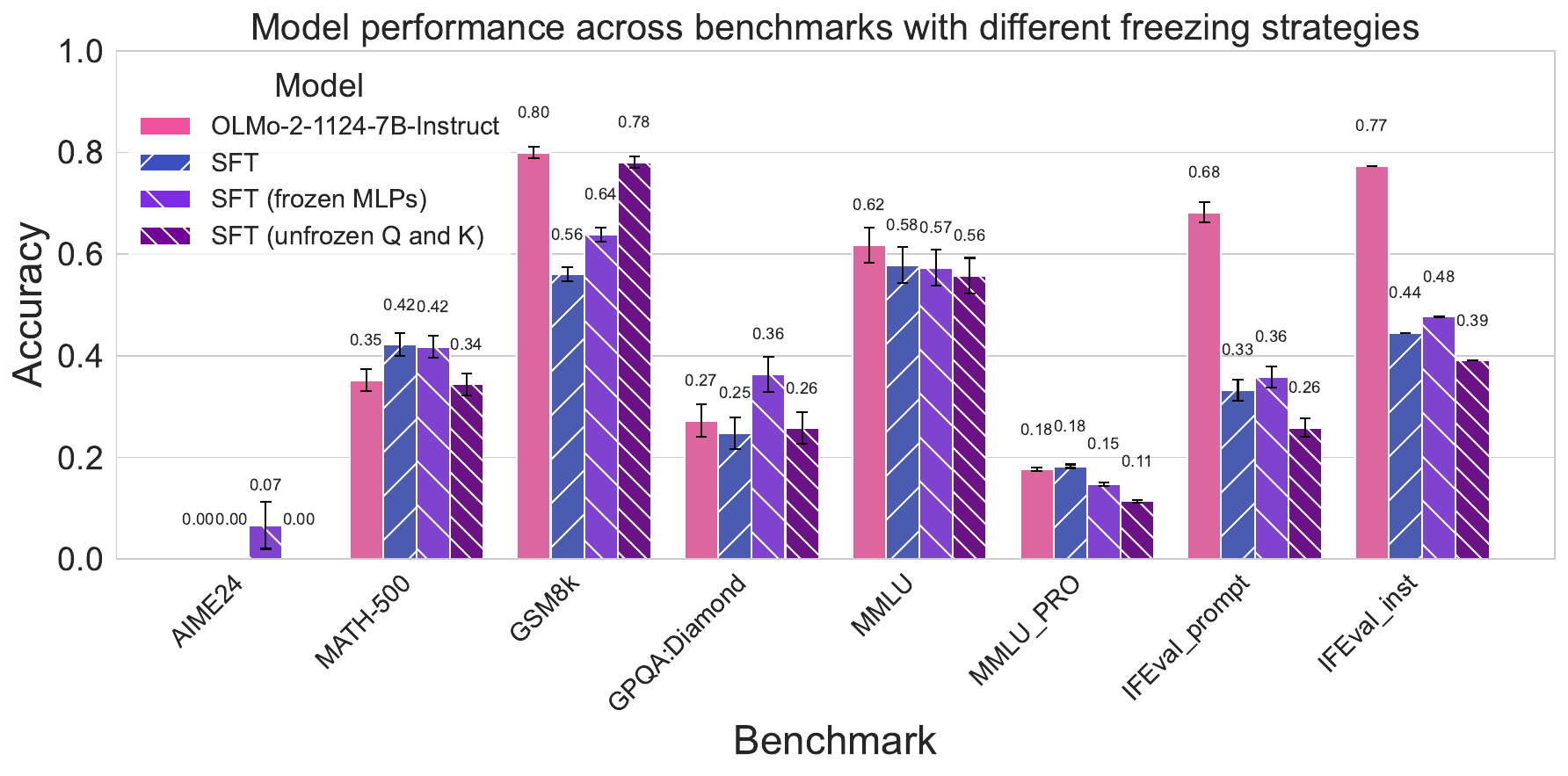}}
\caption{Evaluation results on the same benchmarks as in Figure \ref{fig:benchmarks}. While freezing mid-layer MLPs seems to help on GPQA, it underperforms on other benchmarks. Freezing all parameters aside from Query and Key matrices leads to poor results across the board.}
\label{fig:freezing}
\end{center}
\vskip -0.1in
\end{figure*}

\subsubsection{The parameter-level impact of SFT}
\label{sec:param_impact_sft}
In SFT, we observe that the magnitude of the parameter updates is much greater (note the different scales in the Y axes of Figures \ref{fig:cckpta_grpo} and \ref{fig:cckpta_sft}). This is a plausible reason for the in-domain benefits and out-of-domain degradation observed in Section \ref{sec:evaluating_models_after_reasoning_training}. We also see a similar trend of queries and keys receiving much larger updates than the values and the output. These again seem to be most pronounced in middle layers. Similarly, updates in the MLP matrices are very small at early layers, reaching larger values around the middle layers. Findings by \citet{geva2021transformerfeedforwardlayerskeyvalue} and \citet{meng2023locatingeditingfactualassociations} suggest that mid-layer MLPs are especially important for memorisation and storing factual associations. Therefore, we hypothesise that the larger performance drops in knowledge-intensive domains for SFT in Figure \ref{fig:benchmarks} might stem from these larger mid-layer updates. It is possible that during SFT, circuits or features relating to memorised knowledge contribute highly to incorrect next-token predictions. The loss function might have therefore incentivised overwriting or removing these circuits. This hypothesis could explain why SFT yields larger changes to mid-layer MLPs, corrupting some of that memorised knowledge. Again, our findings are preliminary and inspire further investigations, for example whether freezing these MLPs during SFT can help preserve existing knowledge.

\section{Applying insights from cross-checkpoint analyses via freezing}
In this section, we endeavour to apply the insights we gained in prior sections. Specifically, we investigate whether we can apply our findings on the parameter-level impact of reasoning training practically. In Section \ref{sec:norm_of_differences_throughout_training}, we observed that some model components change notably less than others. This raises the question of whether we can make targeted interventions into reasoning training by freezing parts of the model. We propose two hypotheses informed by our own results as well as prior work. Our results are shown in Figure \ref{fig:freezing}. We experiment with freezing only for SFT to limit computational cost, with every other training detail identical to that during the original SFT runs.

At this point, we emphasize that freezing specific weight matrices in this way may not be particularly principled. Even if freezing mid-layer MLPs could save factual associations, it is unclear whether this will lead to higher performances on knowledge-intensive benchmarks. For one, the training dataset consists of maths questions, so updates to the unfrozen parameters might make the frozen parameters less useful after training. Additionally, \citet{meng2023locatingeditingfactualassociations}'s results on causal tracing might be quite far away from knowledge-intensive tasks as suggested by \citet{hase2023doeslocalizationinformediting}. Therefore, we propose this as a preliminary investigation into making insights from interpretability of reasoning models actionable.

\subsection{Training only the query and key matrices}
\looseness-1
Firstly, we note that the query and the key matrices update the most during both GRPO and SFT. We hypothesise that this is because the primary role of reasoning training is to teach the model to attend to different tokens. For example, the model may be mainly incentivised to refine circuits involving its induction heads \cite{olsson2022context} or iteration heads \cite{cabannes2024iterationheadmechanisticstudy}. Therefore, we investigate whether comparable downstream performance can be attained by only training the query and key matrices of the model.

\looseness-1
As shown in Figure \ref{fig:freezing}, this led to a degradation in performance on MATH-500 below baseline, while performance on GSM8k reduced considerably less than before. Meanwhile, evaluation scores on every other benchmark worsened noticeably. This suggests that training only some parameters significantly limits the model's ability to fit to the training dataset.

\subsection{Freezing the MLP matrices}
\looseness-1
Secondly, we observe that SFT exhibits notable decreases in performance on knowledge-intensive benchmarks. \citet{geva2021transformerfeedforwardlayerskeyvalue} suggest that MLPs act as as key-value memories, and \citet{meng2023locatingeditingfactualassociations} further use causal tracing to indicate that MLPs in mid-layers may be critical for storing factual associations. We replicate their results on OLMo-2-1124-7B-Instruct, noting that it exhibits a similar phenomenon (refer to Appendix \ref{sec:appendix:replicating_romes_results_with_causal_tracing} for further details on causal tracing). Taken together, we ask whether the considerable updates on mid-layer MLPs during SFT might be corrupting factual associations stored in them. To this end, we investigate whether freezing the MLPs highlighted by causal tracing can help retain these associations and prevents degradation of performance on knowledge-intensive benchmarks.


Indeed, as shown in Figure \ref{fig:freezing} we find that freezing the MLPs can have benefits, with performance on MATH-500 remaining stable and degradation on GSM8k compared to full SFT slightly mitigated. Strikingly, evaluation scores on GPQA Diamond improved beyond the base model. While poor results on the remaining benchmarks indicate a nuanced picture, we believe this phenomenon demonstrates early potential for guiding training interventions that are informed by insights from interpretability research. In summary, we find freezing parameters to be neither a reliable way of obtaining the benefits of reasoning training at lower computational cost, nor a way to mitigate loss of factual knowledge.


\section{Conclusion}
\label{sec:conclusion}
In this investigation we present a closer look at the two most widely used methods for training ``reasoning models'', GRPO and SFT. Taking care to eliminate potential confounding variables, we line up both training regimes by training the same model on the same problems with mostly the same hyperparameters. We find GRPO to be expensive and unstable, with easier question subsets and reward function choices being critical for effective training. Meanwhile, we observe SFT to be cheap and dependable. Many works do not account for this cost explicitly, and because SFT is more accessible, this motivates further research into the trade-off. In a controlled comparative setting, we note that GRPO yields modest gains on maths datasets and small degradations outside of them, while both trends are considerably exacerbated by SFT. Looking at the parameter level, we find the updates for query and key matrices to be largest in both methods, but at much larger magnitudes in SFT. We also find greater mid-layer updates in SFT, leading us to hypothesise that this caused the loss in performance on knowledge-intensive tasks. We use these insights to motivate experiments with freezing of model parameters, but find this to work poorly. While our parameter-level analysis reveals notable trends, conflicting results of using them in training shows that further research into this direction is needed. Taken together, our results hint at capability amplification in GRPO, and the acquisition of novel capabilities at the cost of old ones in SFT. Future work could explore the parameter level further, in particular potential sparsity or skewedness of gradient updates.

\section*{Limitations}


While we pursued a fair comparison between both algorithms, we caution the reader that our work comes with drawbacks. For one, OLMo-2-1124-7B-Instruct was already trained with RLVR. While this introduces a confound, we found it to be the best of the available pre- and mid- training variants at actually gaining rewards. Generally, our data setup decisions were made with GRPO due to its cost, but it is possible that SFT could have seen greater gains otherwise. Similarly, it is critical especially for Section \ref{sec:cross-checkpoint_analysis_of_reasoning_training} to ensure that results are outside of the random noise of different seeds. Finally, our method makes a trade-off in keeping almost all hyper-parameters equal. While this makes the training dynamics comparable, it is arguably more realistic to use parameters that allow for optimal performance in each algorithm. We aim to make these improvements in future work. Despite these limitations, we believe that it nonetheless provides a notable initial comparison of GRPO and SFT to build on in future work.


\section*{Acknowledgements}
We express our thanks to the reviewers of this paper for their detailed commentary and for highlighting areas for improvement. We thank the contributors of the open-r1 effort, without which it would have taken us much longer to gain traction on this project. In particular, we thank Edward Beeching for providing advice on hyperparameter configurations for GRPO. We also express our gratitude towards Pasquale Minervini for his support in sourcing access to GPUs as this became necessary. This research was funded by the UKRI AI Centre for Doctoral Training in Responsible and Trustworthy in-the-world Natural Language Processing (grant ref: EP/Y030656/1).



\section*{Impact Statement}
LLMs are being introduced to more and more parts of our lives. Recent progress in their capabilities is remarkable, and it is possible that model developers will continue to advance this frontier. Meanwhile, we understand comparatively little about the internal computations and training dynamics of language models. On the one side, this is irresponsible in and of itself, given how much of an impact LLMs are already having now. On the other side, gaining more of an understanding could potentially lead us to making targeted, internal interventions to alleviate bias or advance safety guardrails. In particular, as paradigms such as ``reasoning training'' are being popularised, it is critical for us to gain a better understanding into what dynamics underpin them. This investigation provides early results into how popular kinds of ``reasoning training'' differ, and what changes they make to models internally. With further follow-up work, we seek to better characterise these dynamics, in hopes of giving recommendations to practitioners on how to keep models aligned and safe while their capabilities advance.

\bibliography{custom}
\bibliographystyle{icml2025}

\newpage
\appendix
\section{Reproducibility}
\label{sec:appendix:reproducibility}
The experiments for this investigation were run on a machine with 8 H100 GPUs via the Edinburgh International Data Facility (EIDF) and the Data-Driven Innovation Programme at the University of Edinburgh. This was necessary due to the computational intensity of GRPO at the 7B scale. For GRPO on smaller models or SFT, much less VRAM is sufficient. All experiments were run via the docker image ``nvcr.io/nvidia/cuda:12.0.0-cudnn8-devel-ubuntu22.04", with NVIDIA driver version 570.124.06 and CUDA version 12.8.

We forked the open-r1 repository\footnote{\url{https://github.com/huggingface/open-r1}} and merged with upstream for critical changes and compute speed-ups. Through the open-r1 repository, we found the highest throughput during GRPO by serving a VLLM server \cite{kwon2023efficient} on one GPU, and launching the actual training job with accelerate \cite{accelerate} on all other GPUs. Under the hood, we found sharding optimizer states and gradients at ZeRO stage-2 with DeepSpeed \cite{rajbhandari2020zeromemoryoptimizationstraining} to be optimal. Our final training configs are shown in Figure \ref{fig:carbon}.

We made significant efforts to align both setups closely to make sure that the impact of each algorithm on model parameters was somewhat comparable. Notably, we aligned all of the data pre-processing and hyperparameters where this made sense. HuggingFace's open-r1 training library \cite{openr1} made this alignment much more feasible, because it builds upon TRL \cite{vonwerra2022trl} for both algorithms. This allows us to keep large parts of the pipeline identical: intricate details like dataset pre-processing or data-parallel model loading and training are all constant between both algorithms. Matching the effective global batch size helps us ensure that the model sees the same question in either approach for a given training step. Furthermore, as mentioned above, in GRPO the model is served on the first GPU, and the other 7 GPUs contribute to the data-parallel global batch size. As such, we also performed SFT with only 7 GPUs in parallel. 

Note that we experimented with many models such as SmolLM2-1.7B-Instruct,  AceInstruct-1.5B, as well as various Qwen models and DeepSeek distils \cite{allal2025smollm2smolgoesbig, liu2025acemathadvancingfrontiermath, Qwen2024, deepseek2025deepseekr1}. However, our final runs use OLMo-2-1124-7B-Instruct \cite{olmo20252olmo2furious}. For its size, it is the most capable model with fully open pre-, mid-, and post-training data as well as training code, intermediate checkpoints, and tools such as OLMoTrace . This enables future investigations into which capabilities it has before our training, which ones are reinforced and when it originally acquired them.

Unfortunately, OLMo-2-1124-7B-Instruct supports a maximum of 4096 tokens. As such, we filtered out all training examples where a long reasoning trace by R1 would result in more than 4096 tokens. To this end, we pre-processed the dataset by applying the chat template and the system prompt to every problem/solution pair, tokenizing this and removing the sample if it exceeded 4096 tokens. This ensured that during SFT, we only trained on problems where the closing \textless /think\textgreater and the contents of \textless answer\textgreater \textless /answer\textgreater were included. We found performance to suffer otherwise, likely because the model would sometimes run out of tokens in its chain-of-thought. We use the same dataset for training after this pre-processing in both algorithms.

While this allowed us to mitigate some issues arising from 4096-token length limit, others were harder to circumvent. Notably, setting a maximum completion length of 4096 during GRPO would cause out-of-memory errors whenever copies of the model starting using up that token budget while attempting hard problems. As such, we were forced to set the maximum completion lengths to 1500 in GRPO. This is clearly a confound, as completions which are uncompleted get masked during our training setup, and in GRPO we perform backward passes with respect to fewer tokens. Further, the model necessarily will learn to generate fewer tokens during GRPO. We consider mitigating this problem an important area for future work, for example by reducing the batch size per device but increasing the gradient accumulation steps. This could reduce memory usage while keeping the global batch size the same, allowing for higher maximum completion lengths.

Finally, note that the open-r1 repository also features support for GRPO on maths and code in parallel, closely following the approach used by \citet{deepseek2025deepseekr1}. We focus on maths questions only, leaving comparisons of multi-domain reasoning training for future work.


\section{Finding working GRPO configurations for OLMo-2}
\label{sec:appendix:finding_working_GRPO_configrations_for_olmo-2}
Here we outline a number of lessons we learned while tweaking our GRPO setup to achieve successful reasoning training with OLMo-2.

\subsection{Question difficulty}
\label{sec:appendix:question_difficulty}

\begin{figure}[h]
\begin{center}
\centerline{
\includegraphics[width=0.9\columnwidth]{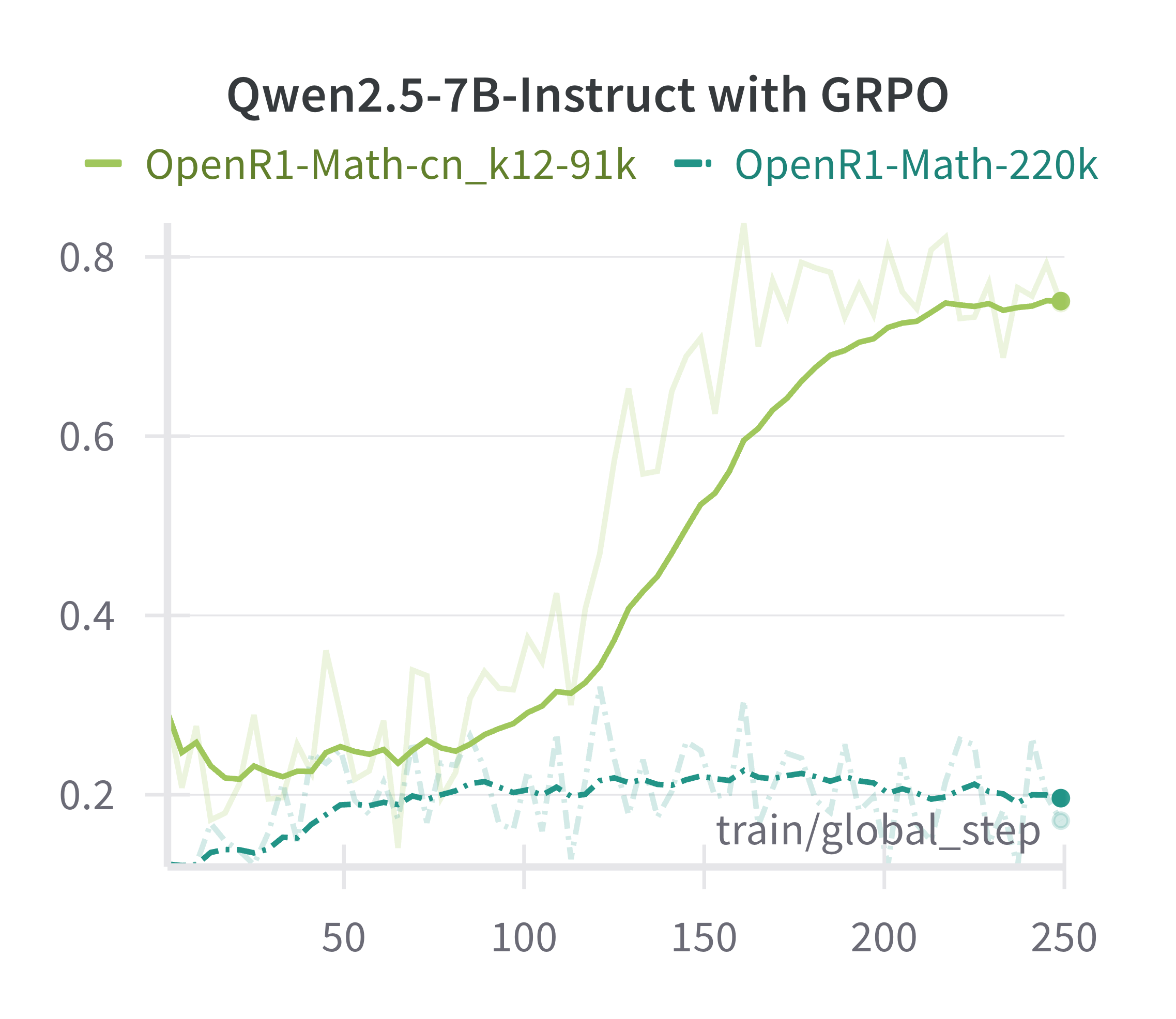}
}
\caption{Accuracy rewards of Qwen2.5-7B-Instruct during GRPO on questions sampled from OpenR1-Math-220k and the CN-K12 subset. In the former, the model continuously struggles on many of the questions, while in the latter, the model gradually exhibits much stronger performance. Note that both curves display stronger performance than subsequent Figures based on OLMo-2-1124-Instruct. We started with Qwen to find any recipe that worked, and then adapted it to OLMo.}
\label{fig:Math-220kvsMathcnk12-91k}
\end{center}
\vskip -0.1in
\end{figure}

We find the largest factor in training success to be the relationship between base model capability and question difficulty. The default training split of OpenR1-Math-220k contains 93k questions, of which 72.6\% stem from olympiad competitions. In preliminary trials with SmolLM2-1.7B-Instruct and AceInstruct-1.5B \cite{allal2025smollm2smolgoesbig, liu2025acemathadvancingfrontiermath}, $\leq2$B models generated incorrect responses to these questions so often that they never improved. Scaling to 7B yielded improved training stability and accuracy rewards at much greater computational costs, in line with previous findings in the literature \cite{gurung2025learningreasonlongformstory}. However, even 7B models still struggled with many of the olympiad level questions. As such, we instead use the extended training split of OpenR1-Math-220k, which consists to 69.6\% of much easier CN-K12 questions. We further filter the dataset to contain only these 91k CN-K12 questions, observing a notable improvement in training effectiveness. Figure \ref{fig:Math-220kvsMathcnk12-91k} shows the accuracy rewards of Qwen2.5-7B-Instruct during an experiment in an early phase of our investigation. We find it to receive increasing accuracy rewards when training on the CN-K12 subset much more reliably than on the full dataset, with the trend even more pronounced in OLMo-2-1124-7B-Instruct. No other adjustment to the training setup has as pronounced of an impact as changing the question difficulty.

This ties in with previously mentioned recent findings in the literature \cite{liu2025understandingr1zeroliketrainingcritical, zhao2025echochamberrlposttraining, hochlehnert2025soberlookprogresslanguage, gandhi2025cognitivebehaviorsenableselfimproving}. It confirms that GRPO reinforces capabilities the model already had, and that it does not help the model learn novel tasks. We also qualitatively observe that OpenR1-Math-220k questions which the models get right are easier. This connects in particular with results presented by \citet{yue2025doesreinforcementlearningreally}. They suggest that RLVR shifts the output distribution towards generating the correct response at pass@1, but only when the base model was already capable of generating the correct response in pass@\textit{k} for high values of \textit{k}. \citet{tang2025optimizinglanguagemodelsinference} argue that this is not a bug but a feature of GRPO in this setup. As our focus lies in comparing GRPO to SFT, we leave this question for other work. Our priority is a GRPO configuration where the model receives rewards consistently and improves performance to compare the underlying parameter updates. 

\subsection{Reward function design}
\label{sec:appendix:reward_function_design}

\begin{figure}[h]
\begin{center}
\centerline{\includegraphics[width=0.9\columnwidth]{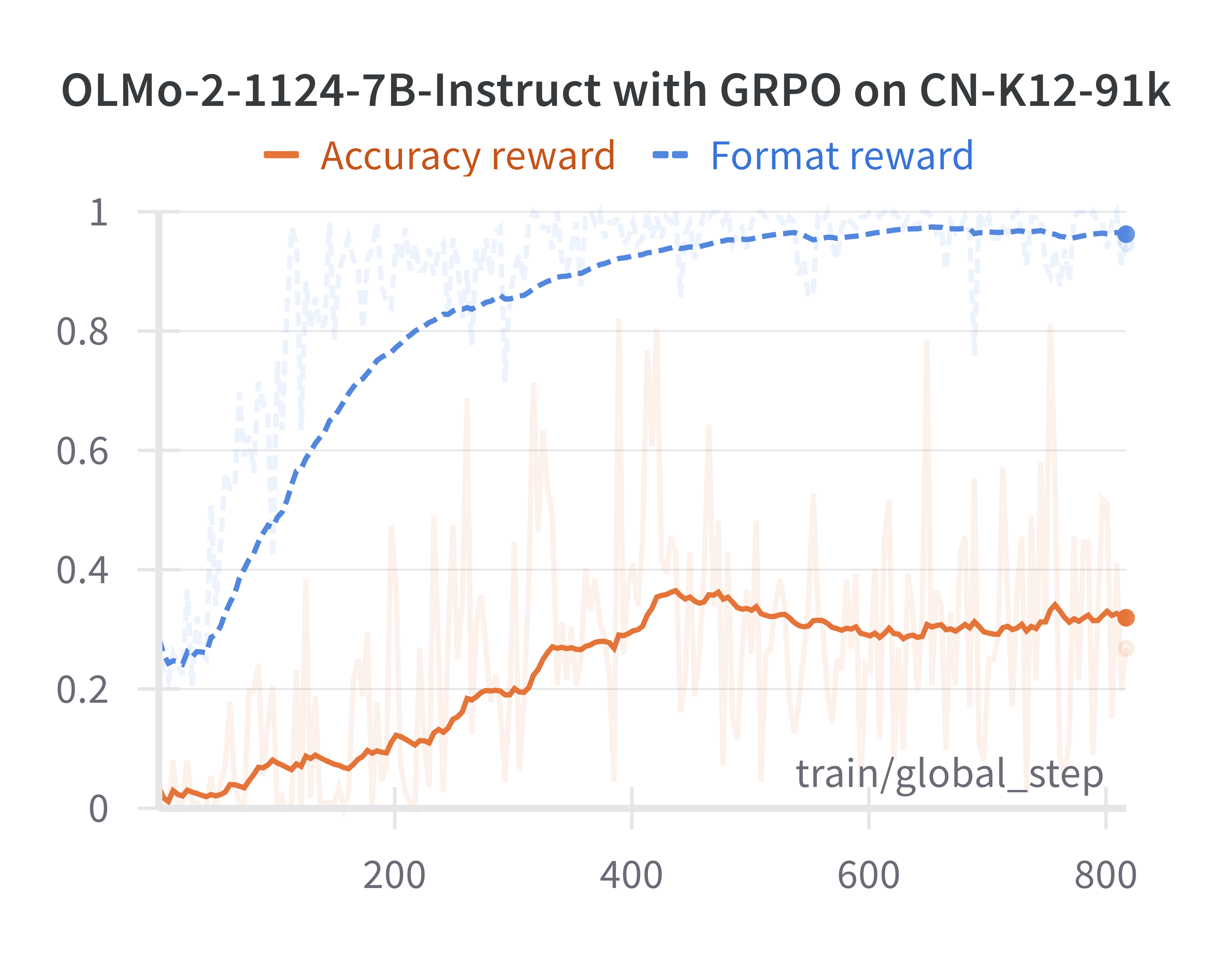}}
\caption{Accuracy and format rewards of OLMo-2-1124-Instruct during GRPO on questions sampled from CN-K12-91k subset. Format rewards increase quickly as the model follows instructions correctly, while accuracy rewards increase more slowly and plateau.}
\label{fig:format_accuracy_appendix}
\end{center}
\vskip -0.1in
\end{figure}

We find another crucial factor in successful GRPO training to be correct reward function design. Out of the box, the open-r1 library enables three reward functions: 
\vspace{-0.5em}
\begin{itemize}
\setlength{\itemsep}{0pt}
\setlength{\parskip}{0pt}
\setlength{\parsep}{0pt}
\item to reward rollouts from which correct solutions were successfully extracted
\item to check whether the preliminary reasoning was enclosed in the \textless think\textgreater tags
\item to add 0.25 to the reward for every one of these four XML tags asked for in the system prompt
\end{itemize}
\vspace{-0.5em}
We observe a very intricate interplay between these three reward sources. The accuracy rewards seem the most decisive, as we notice many instances where the model generates a correct response expressed differently than the solution in the dataset. Actually rewarding instances where a correct answer is extracted from a response therefore seems crucial to update parameters towards successful reasoning. 

Meanwhile, for format and tag count rewards, the picture is a bit more nuanced. On the one hand, incentivising model responses that follow the \textless think\textgreater  instructions faithfully appears important to promote step-by-step reasoning before generating an answer. On the other hand, there are many questions where the model generates incorrect responses. 

Continuously updating model parameters with respect to incorrect reasoning traces because they follow the right format is undesirable. This is especially true for long training runs where the model learns the formatting quickly, but struggles to improve its training accuracy. Because the advantage calculation in GRPO is relative to the mean and standard deviation of the reward in the group, this effect is mitigated somewhat. But we find that in practice, there is a significant number of completions with incorrect solutions that still lead to parameter updates due to advantages on the format rewards. An instance of this leading to instability is shown in Appendix \ref{sec:appendix:GRPO_failures}, with exploding gradient norms and a specific prompt-completion pair during training shown in Figures \ref{fig:grad_norm} and \ref{fig:xml_tags}. Following continuous updates with high format reward, the model abruptly starts to end responses with random XML tags.

Therefore, we choose to omit tag count rewards, and scale accuracy rewards and format rewards by 1.0 and 0.1, respectively. We find this to be an effective compromise between teaching the model to follow the system prompt quickly, while placing much greater emphasis on rollouts that led to correct solutions. Figure \ref{fig:format_accuracy_appendix} shows how the model quickly learns to follow formatting instructions, and accuracy reward increases gradually before plateauing. Another simple solution would be to only grant format rewards to a completion if the answer was correct, but exploring this further is out of scope for our investigation.

\subsection{Training stability}
\label{sec:appendix:training_stability}

Finally, we find tweaking hyperparameters for stability to be very impactful for training effectiveness. In particular, we focus on maximising effective global batch sizes without running out of memory in a data parallel setup. When scaling batch sizes with TRL, there is a specific dynamic between the batch size per device, gradient accumulation steps, and the number of rollouts per individual prompt\footnote{See \url{https://github.com/huggingface/trl/blob/v0.17-release/trl/trainer/grpo_trainer.py\#L733}}.

We synchronise our configurations between SFT and GRPO and optimise for maximum throughput and effective batch sizes without running out of memory. Our most consistently stable results are at a global batch size of 112, with 28 completions generated per prompt, and 1 gradient update per global step. This improved the reliability of training noticeably, and alleviated issues such as in Figures \ref{fig:grad_norm}, \ref{fig:xml_tags} and \ref{fig:kl_explosion}. For the full set of hyperparameters used for training, refer to Appendix \ref{sec:appendix:reproducibility}.

\begin{figure*}[hb]
\centerline{\includegraphics[height=0.95\textheight]{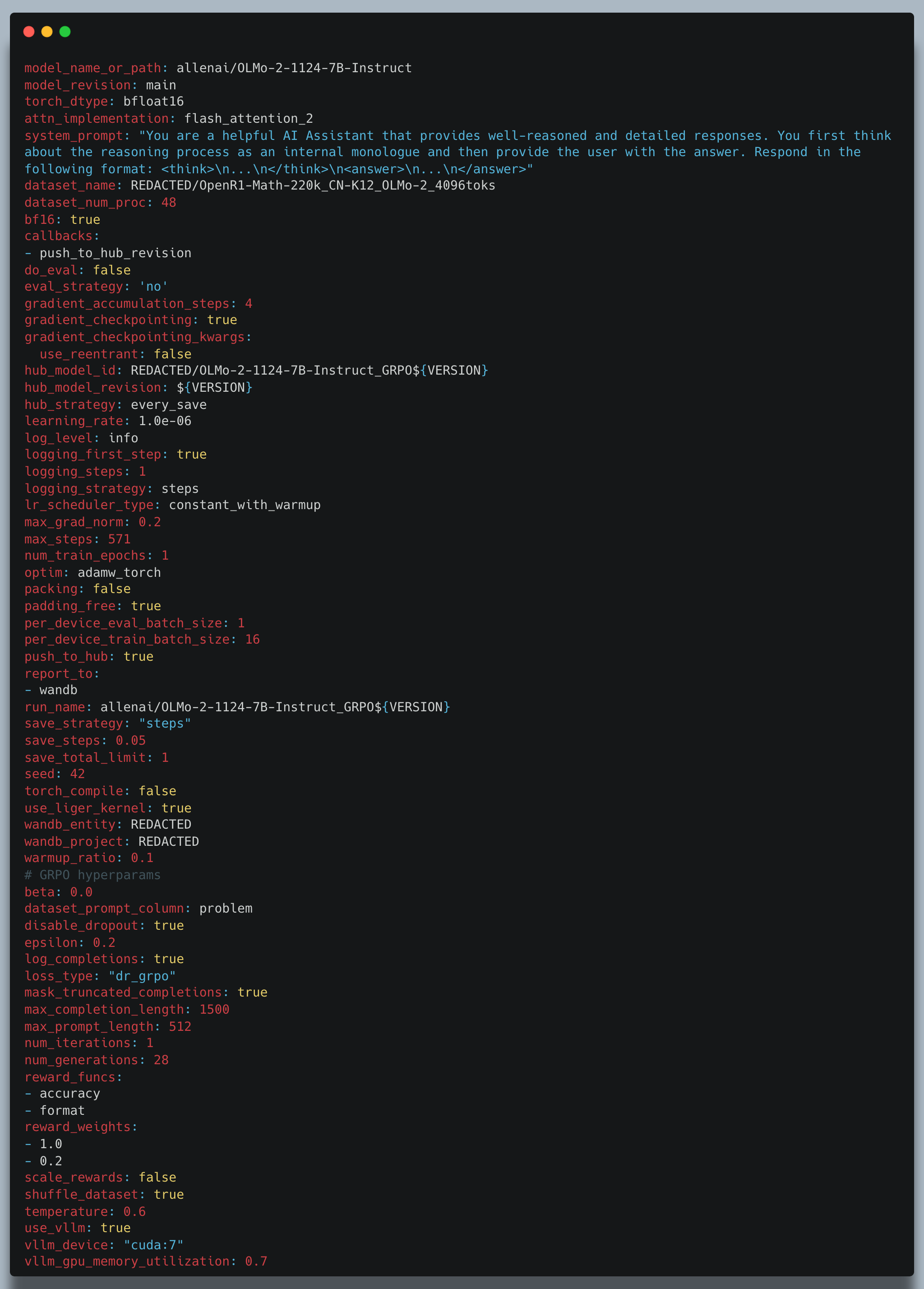}}
\caption{Our training parameters. SFT parameters were identical, except for a learning\_rate of 5e-5 and the GRPO parameters removed.}
\label{fig:carbon}
\end{figure*}

Note that \citet{shao2024deepseekmathpushinglimitsmathematical} actually perform multiple gradient updates per global step (this corresponds to the hyperparameter $\mu$ in their original GRPO algorithm, and "num\_iterations" in the TRL config in Figure \ref{fig:carbon}). However, when increasing it to a value such as 4, TRL consumes the total number of steps 4 times as quickly\footnote{Refer to \url{https://github.com/huggingface/trl/pull/2899} for more details.}. This would mean that the model is trained on a different number of problems during training. We consider this to be too big of a confounding variable where training under GRPO, and as such keep it fixed at 1.

Finally, we make use of insights from very recent work like Dr GRPO \cite{liu2025understandingr1zeroliketrainingcritical}, DAPO \cite{yu2025dapoopensourcellmreinforcement} and blogs by the open-r1 team \cite{openr1update2}. From these we integrate adjustments such as removing the length bias of GRPO and incorporating various adjustments as the open-r1 and TRL repositories are improved. We replicate their findings that completion lengths remain stable throughout training, though it is possible that this stems from our exclusion of harder olympiad questions. Our evaluations on a diverse set of benchmarks are detailed in Section \ref{sec:evaluating_models_after_reasoning_training}.

\section{Unsuccessful experiments with GRPO}
\label{sec:appendix:GRPO_failures}

\begin{figure*}[h]
\begin{center}
\centerline{\includegraphics[width=0.9\textwidth]{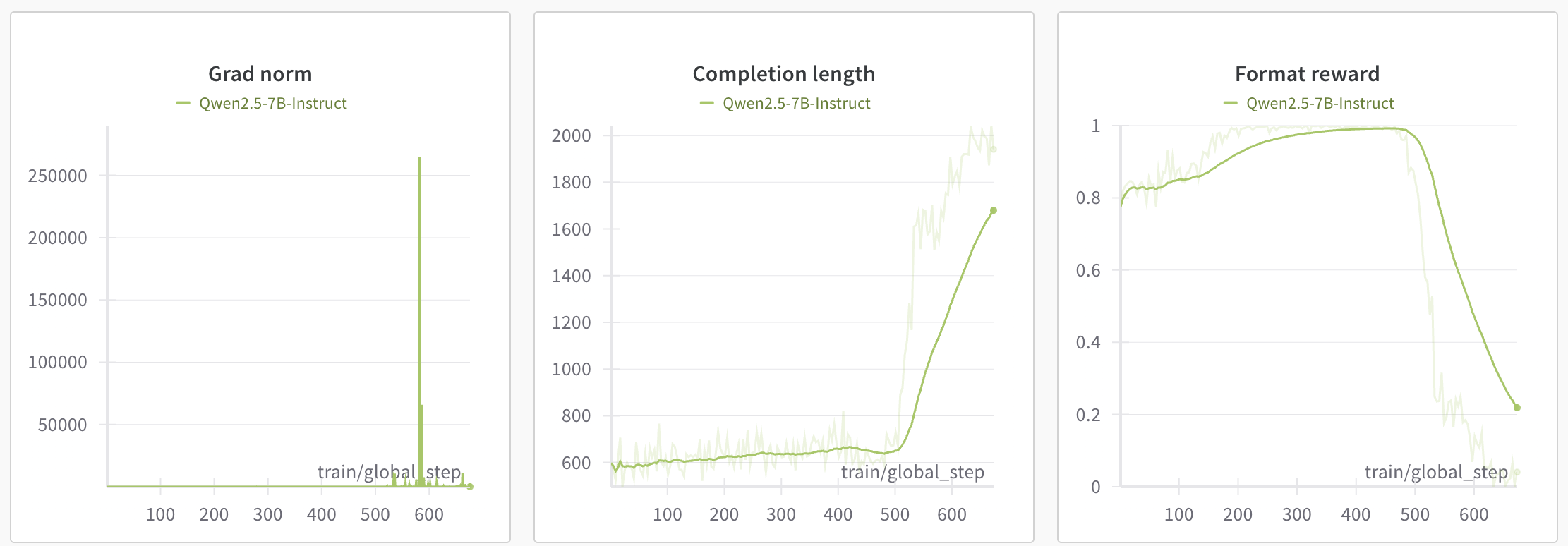}}
\caption{Grad norm of updates, format rewards and completion lengths of Qwen-2.5-7B-Instruct during GRPO on questions from CN-K12-91k subset. After successfully learning the format, the model starts spamming XML tags until reaching the maximum completion length of 2048, with explosions in the grad norm indicating training instability.}
\label{fig:grad_norm}
\end{center}
\vskip -0.1in
\end{figure*}

\begin{figure*}[h]
\begin{center}
\centerline{\includegraphics[width=\textwidth]{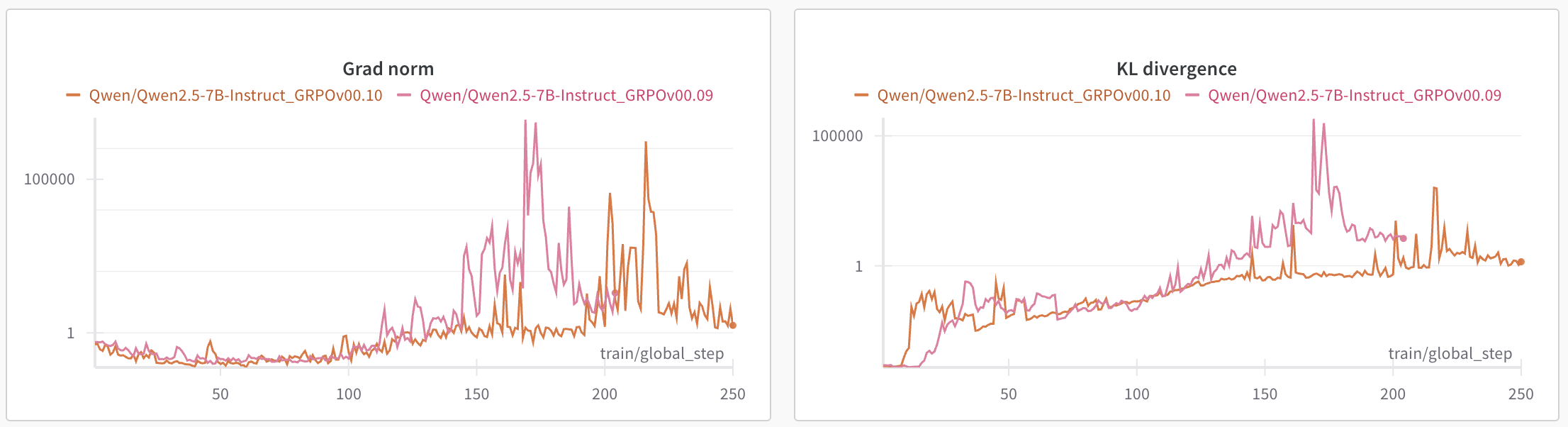}}
\caption{Grad norm of updates and KL divergence between the current policy and reference policy (ie. Qwen-2.5-7B-Instruct before and during during GRPO) on questions from CN-K12-91k subset. Note the log scale on the y-axis. Before finding hyperparameters with greater training stability (especially larger global effective batch sizes), training runs were often unreliable.}
\label{fig:kl_explosion}
\end{center}
\vskip -0.1in
\end{figure*}

\begin{figure*}[h]
\begin{center}
\centerline{\includegraphics[width=\textwidth]{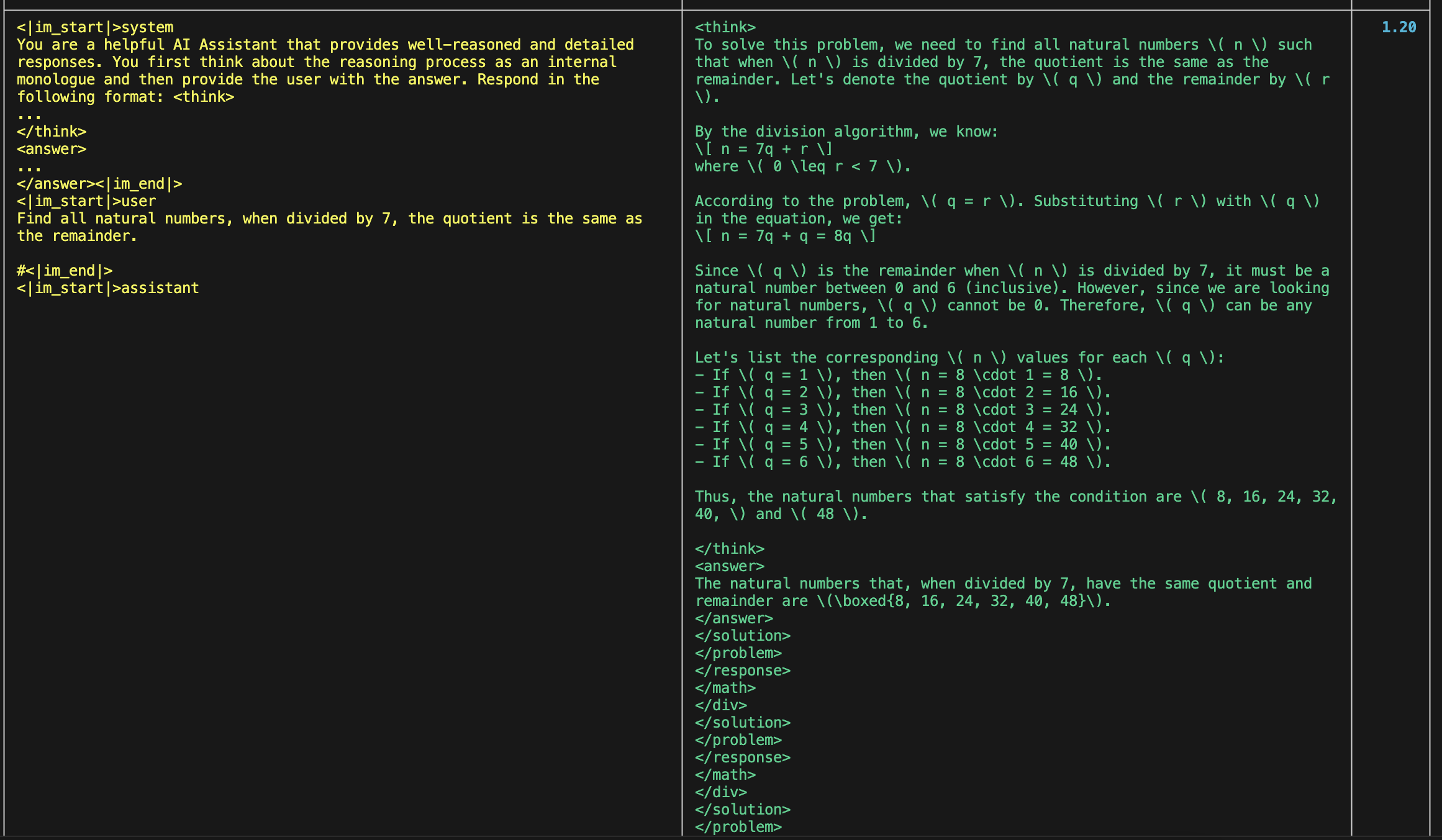}}
\caption{Screenshot of the CLI during the unstable training run shown in Figure \ref{fig:grad_norm}. The full prompt, including the system prompt and user instruction, is on the left, while the model completion is in the middle. The model follows the formatting as instructed, but after the final \textless /answer\textgreater it continues to spam hallucinated XML tags instead of the EOS token. Notably, the model response is correct, and as visible in the right column, it received full reward (the accuracy and format reward weightings were 1.0 and 0.2 for this run).}
\label{fig:xml_tags}
\end{center}
\vskip -0.1in
\end{figure*}

We tried a number of different approaches to GRPO that did not work. As mentioned above, open models smaller than 7B, datasets with hard questions, and poorly adjusted reward functions were all sources of unstable training. This is particularly evident in the ``reward hacking'' case shown in Figures \ref{fig:grad_norm} and \ref{fig:xml_tags}. Furthermore, we found learning rates higher than 1e-6 to be very unreliable, even crashing after the accuracy rewards and format rewards had been stable for 100s of steps, or leading to huge updates or divergences as shown in Figure \ref{fig:kl_explosion}.

\begin{figure}[h]
\begin{center}
\centerline{\includegraphics[width=\columnwidth]{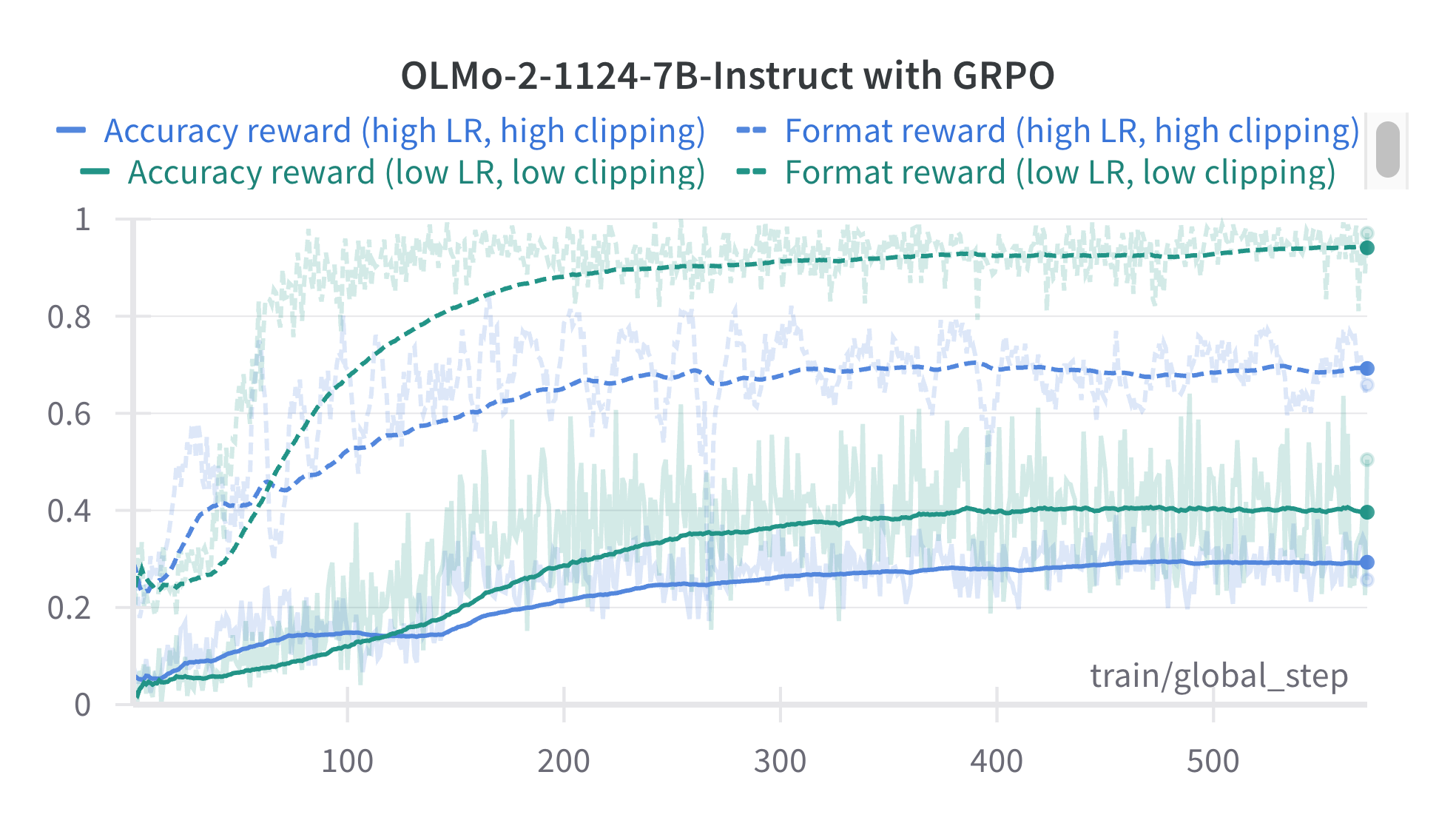}}
\caption{Accuracy and format rewards when using low learning rates and clipping (baseline), or high learning rates and clipping.}
\label{fig:high_lr_high_clipping}
\end{center}
\vskip -0.1in
\end{figure}

At the same time, SFT consistently achieved better performances at a learning rate around 5e-5. In an attempt to align the learning rate for both algorithms, we attempted to raise it in GRPO from 1e-6 to 1e-5. However, accuracy rewards would consistently crash after around 100 steps. As such, we tried to pair it with stricter regularisation. Specifically, we combined the learning rate of 1e-5 with a reduced max\_grad\_norm from 0.2 to 0.05, while re-introducing the KL Divergence loss term with the scaling term $\beta$ at 0.04, and tighter clipping with the scaling term $\epsilon$ reduced from 0.2 to 0.1. Figure \ref{fig:high_lr_high_clipping} compares the performance of this to our baseline results (at a learning rate of 1e-6 and less clipping). While all of the regularisation parameters enabled slightly more stable training, it also resulted in consistently worse evaluation results on every benchmark. As such, we abandoned this line of investigation.

We noticed that another important component of the pipeline was the benchmark setup. For example, we were inadvertently training and evaluating at different temperatures initially, leading to lower benchmark scores. Similarly, we found that only setting the ``system\_prompt'' parameter in the open-r1 YAML file did not actually modify the system prompt of the model saved locally or pushed to HuggingFace at the end of training. That meant that during evaluation after training, the model was using the default OLMo-2 chat template instead of the DeepSeek-style ``thinking'' template used during training. Fixing these discrepancies enabled much more consistent results, in line with findings by \citet{hochlehnert2025soberlookprogresslanguage}.

\section{Revisiting Causal Tracing with OLMo-2-1124-7B-Instruct}
\label{sec:appendix:replicating_romes_results_with_causal_tracing}
\begin{figure}[h]
\begin{center}
\centerline{\includegraphics[width=\columnwidth]{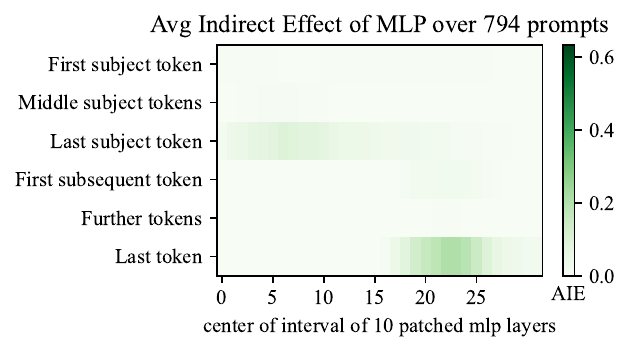}}
\caption{The average indirect effect (AIE) of the MLPs of OLMo-2-1124-7B-Instruct. We use the same exact hyperparameters and dataset as \citet{meng2023locatingeditingfactualassociations}, though not all of the \(\sim \)1200 prompts were answered correctly by OLMo-2-1124-7B-Instruct, and so only 794 are used for computing AIE.}
\label{fig:rome}
\end{center}
\vskip -0.1in
\end{figure}

In order to help motivate our experiments on freezing mid-layer MLPs, we turn towards prior findings by \citet{meng2023locatingeditingfactualassociations}. In particular, we make use of their publicly available repository and dataset\footnote{\url{https://github.com/kmeng01/rome}}, and adapt them to add support for OLMo-2-1124-7B-Instruct. Our results are shown in Figure \ref{fig:rome}.

We find OLMo-2-1124-7B-Instruct to exhibit similar phenomena as the models examined by \citet{meng2023locatingeditingfactualassociations}. There seems to be an 'early site' of MLPs that are causally implicated on the last subject token, and a 'late site' of MLPs in upper layers with high average indirect effect (AIE) on the last token. We use these results to help motivate which layers to freeze. Specifically, due to the apparent relative strength of the 'late site', we choose to freeze MLPs in all layers where AIE is greater than 0.1, namely layers 20-26. Our results are shown in Figure \ref{fig:freezing}.

\label{sec:GRPO_hyperparams}
\vspace{-0.1in}
\vspace{-0.1in}

\clearpage


\end{document}